\documentclass[conference]{IEEEtran}
\IEEEoverridecommandlockouts
\usepackage{cite}
\usepackage{amsmath,amssymb,amsfonts}
\usepackage{graphicx}
\usepackage{textcomp}
\usepackage{xcolor}
\usepackage{subfigure}
\usepackage{bm}
\usepackage{booktabs}
\usepackage{multirow}
\usepackage{authblk}
\usepackage{algorithm}  
\usepackage{algorithmicx}  
\usepackage{algpseudocode}
\usepackage{booktabs}
\usepackage{threeparttable}

\usepackage{pifont}
\usepackage{pgfplots}

\usepackage{comment}
\usepackage[switch]{lineno}
\newcommand{\red}[1]{\textcolor{red}{#1}}

    \def\addlegendimage{\csname pgfplots@addlegendimage\endcsname}

\definecolor{pt1min}{HTML}{1A237E}
\definecolor{pt2min}{HTML}{B71C1C}
\definecolor{pt4min}{HTML}{1B5E20}
\definecolor{wopt1min}{HTML}{01579B}
\definecolor{wopt2min}{HTML}{BF360C}
\definecolor{wopt4min}{HTML}{004D40}
\definecolor{pttte}{HTML}{E65100}
\definecolor{wopttte}{HTML}{FFD600}

\def\BibTeX{{\rm B\kern-.05em{\sc i\kern-.025em b}\kern-.08em
    T\kern-.1667em\lower.7ex\hbox{E}\kern-.125emX}}

\begin{document}



\title{DRL4AOI: A DRL Framework for Semantic-aware AOI Segmentation in Location-Based Services}


\author{
	Youfang Lin\textsuperscript{\rm 1, 2},
        Jinji Fu\textsuperscript{\rm 1},
        Haomin Wen\textsuperscript{\rm 1,3},
        Jiyuan Wang\textsuperscript{\rm 1}, 
        Zhenjie Wei\textsuperscript{\rm 1},
        Yuting Qiang\textsuperscript{\rm 3}, \\
        Xiaowei Mao\textsuperscript{\rm 1, 2},  
        Lixia Wu\textsuperscript{\rm 3},
        Haoyuan Hu\textsuperscript{\rm 3},
        Yuxuan Liang\textsuperscript{\rm 4},
	Huaiyu Wan\textsuperscript{\rm 1, 2, *}, \\
	\thanks{~~Corresponding author: hywan@bjtu.edu.cn}
	\textsuperscript{\rm 1}School of Computer and Information Technology, Beijing Jiaotong University, Beijing, China\\
	\textsuperscript{\rm 2}Beijing Key Laboratory of Traffic Data Analysis and Mining, Beijing, China\\
        \textsuperscript{\rm 3}Cainiao Network, Hangzhou, China. \\
        \textsuperscript{\rm 4}Hong Kong University of Science and Technology (Guangzhou) \\

	\{yflin, wenhaomin, maoxiaowei, 20271271, 20271161, 20241068, hywan\}@bjtu.edu.cn,\\
	\{xiushui.qyt,wallace.wulx,haoyuan.huhy\}@cainiao.com; yuxliang@outlook.com
	
}


\maketitle

\begin{abstract}

In Location-Based Services (LBS), such as food delivery, a fundamental task is segmenting Areas of Interest (AOIs), aiming at partitioning the urban geographical spaces into non-overlapping regions. Traditional AOI segmentation algorithms primarily rely on road networks to partition urban areas. While promising in modeling the geo-semantics, road network-based models overlooked the service-semantic goals (e.g., workload equality) in LBS service. In this paper, we point out that the AOI segmentation problem can be naturally formulated as a Markov Decision Process (MDP), which gradually chooses a nearby AOI for each grid in the current AOI's border. Based on the MDP, we present the first attempt to generalize Deep Reinforcement Learning (DRL) for AOI segmentation, leading to a novel DRL-based framework called DRL4AOI. The DRL4AOI framework introduces different service-semantic goals in a flexible way by treating them as rewards that guide the AOI generation. To evaluate the effectiveness of  DRL4AOI, we develop and release an AOI segmentation system. We also present a representative implementation of DRL4AOI — TrajRL4AOI — for AOI segmentation in the logistics service. It introduces a Double Deep Q-learning Network (DDQN) to gradually optimize the AOI generation for two specific semantic goals: i) trajectory modularity, i.e., maximize tightness of the trajectory connections within an AOI and the sparsity of connections between AOIs,  ii) matchness with the road network, i.e., maximizing the matchness between AOIs and the road network. Quantitative and qualitative experiments conducted on synthetic and real-world data demonstrate the effectiveness and superiority of our method. The code and system is publicly available at https://github.com/Kogler7/AoiOpt.



\end{abstract}



\section{Introduction}
Location-based services (such as food delivery, logistics, ride-sharing and spatial crowdsourcing\cite{zhao2020predictive, zhao2019destination}) have experienced rapid growth by greatly facilitating people's lives. One fundamental spatiotemporal data management \cite{Chen2023kdd} behind these LBS platforms is partitioning urban geographical space into multiple small, non-overlapping Areas of Interest (AOIs). For instance, ride-sharing giants like DiDi\footnote{https://web.didiglobal.com/} and Uber\footnote{https://www.uber.com/} divide the city into several AOIs to dispatch idle drivers to high-demand areas \cite{geng2019spatiotemporal, liu2022deep}. Similarly, logistics platforms like Cainiao\footnote{https://global.cainiao.com/} and JD.COM\footnote{https://global.jd.com/}  dispatch orders to couriers based on AOI segmentation \cite{wu2023lade, drl4route2023mao, ruan2022discovering}. The quality of the AOI segmentation directly influences the service quality. In light of the above cases, there is a rising call for effective AOI segmentation methods to generate a well-managed set of AOIs.

\begin{figure}[!t]
    \centering
    \includegraphics[width=1.1 \linewidth]{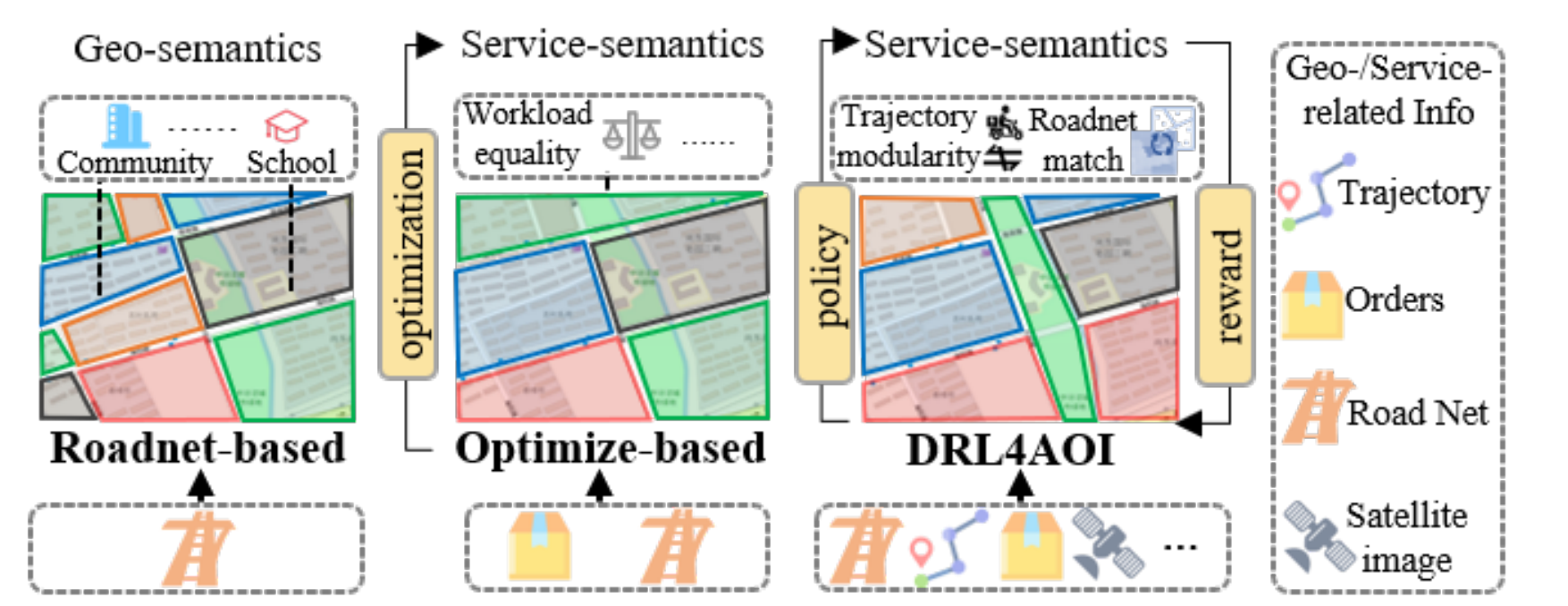}
    \caption{Illustration of Different Methods.}
    \label{fig:introduction}
    \vspace{-1em}
\end{figure}

\par As shown in Figure~\ref{fig:introduction}, an intuitive way is road network-based methods, which divide AOIs according to road networks \cite{muyldermans2002districting, lei2012districting}. Though promising in preserving the geo-semantics (such as identifying a geographical entity, i.e., school), road network-based models are not designed to meet various service-semantics goals  (e.g., workload equity \cite{guo2023towards}) in LBS. To this end, there is a rising trend of optimization-based approaches that aim to meet the service-semantic goals. For instance, E-partition\cite{guo2023towards} segments AOIs with the goal of workload equity, and RegionGen\cite{Chen2023kdd} aims to achieve more accurate demand prediction when generating AOIs.  Though promising, those optimization-based algorithms are challenged to model abundant spatial and temporal features in the optimization process, which restricts their performance in real-life scenarios.


\begin{table}[htbp]
	    \centering
	    \caption{Comparison between our model and related ones.}
    	\resizebox{\linewidth}{!}{
    		\begin{tabular}{cccccc}
    			\toprule
    			Method & Model geo-semantic?  & Model service-semantic? &  Abundant features?\\
    			\midrule
    			  Fixed-shape & \ding{53} & \ding{53} & \ding{53}\\
    			Road-network-based & \ding{52} & \ding{53} & \ding{53}\\
                    Optimization-based & \ding{52} & \ding{52} & \ding{53}\\
    			\midrule
    			\midrule
    			 DRL4AOI (ours)  & \textbf{\ding{52}} & \textbf{\ding{52}}  & \textbf{\ding{52}} \\
    			\bottomrule
    		\end{tabular}
    	}
	\label{tab:method_compare}
\end{table}




\par To address the above limitation, we resort to Deep Reinforcement Learning (DRL) for the AOI segmentation, for it combines the power of RL methods in non-differentiable objective optimization with deep learning models in complex feature learning. Specifically, we provide a novel perspective that shows that the AOI segmentation can be naturally formulated as a Markov Decision Process (MDP), where the MDP gradually chooses a nearby AOI for each grid in the current AOI's border. Following that, we present the first attempt to generalize Deep Reinforcement Learning (DRL) for the problem, leading to the first-ever DRL-based AOI segmentation framework called DRL4AOI. It introduces different service-semantic goals in a flexible way by treating them as rewards that guide the AOI generation. Table~\ref{tab:method_compare} lists the comparison of our method and previous efforts.

\par Moreover, based on the framework, we further propose a model named TrajRL4AOI for AOI segmentation in the logistics service. It aims to achieve two kinds of service-semantic goals: i) trajectory modularity, i.e., maximize tightness of the trajectory connections between nodes within an AOI, and the sparsity of connections between AOIs, ii) matchness with road network, i.e., maximizing the matchness between AOIs and the road network. In TrajRL4AOI, a Deep Q-learning Network (DQN) is devised to optimize the AOI generation during training. Overall, we summarize our contribution as the following three points:



\begin{itemize}
    \item We hit the problem of the AOI segmentation task from a reinforcement learning perspective with the first shot, leading to a semantic-aware DRL-based framework called DRL4AOI. It combines the power of RL methods in service-semantic goal optimization with deep learning models in abundant feature modeling.
   
    \item Based on the framework, we propose TrajRL4AOI for AOI segmentation in logistics service. By treating the AOI segmentation as an MDP that makes AOI decisions for each grid on AOI's border, it utilizes a value-based model, Double-DQN, to improve the trajectory modularity and road-network matchness with the road network.
    

    \item We develop and release an AOI segmentation system, and extensive offline experiments conducted on synthetic data and real-world data demonstrate the superiority of our method over other solutions.
    

\end{itemize}

\section{Related Work} \label{sec:related_work}

\par Current AOI segmentation methods can be broadly classified into three classes: 1) fixed-shape, 2) road-network-based, and 3) optimization-based methods.

\par \textbf{Fixed-shape methods}. Fixed-shape methods divide the urban space into several fixed-shape grids or hexagons \cite{ke2017short, geng2019spatiotemporal, ling2023sthan, zhang2017deep}. It is a simple and effective way to output a partition that naturally covers the city, i.e., any point in the target space is associated with an AOI. However, those methods cannot capture the geographic semantic meaning of the urban space, which may significantly trim down the performance of other AOI-based tasks in the platform.

\par \textbf{Road-network-based methods}. Targeting at solving the shortcomings of fixed-shape methods, road-network-based methods \cite{muyldermans2002districting, lei2012districting} capture the geographic semantics by dividing the urban space via road networks. In this way, nearby locations with the same geographic meaning (e.g., all areas in a school) can be grouped into one AOI. Though effective in most scenarios, a fundamental limitation of the road-network-based method \cite{Chen2023kdd, guo2023towards} is the lack of service semantic information,  which is usually considered a bottleneck that restricts the overall system performance.



\par \textbf{Optimization-based methods.} To address the above limitation, optimization-based methods have emerged in recent years. Those models introduce service-specific goals or constraints (i.e., service semantics) to guide the AOI generation based on optimization-based methods by levering various kinds of data generated in the service. For example, \cite{winkenbach2016enabling,bender2020districting,zhou2021heuristic,banerjee2022fleet}, combines downstream route planning tasks and establishes a cost function based on factors like workload and distance. They employ heuristic search algorithms to solve the optimization problem and partition the regions accordingly. RegionGen \cite{Chen2023kdd} models the AOI generation as a multi-objective optimization problem. It first segments the city into atomic spatial elements based on road networks. Then, it clusters the above elements into different AOIs by maximizing various operation goals, i.e., clusters' average predictability and service specificity. \cite{arampatzis2006web} introduced an alpha-shape partitioning method based on POI data, extracting relevant data from the Web and using the alpha-shape method to infer the boundaries of imprecise AOIs.  C-AOI \cite{zhu2023c} propose an image-based model that converts the AOI segmentation into an instance segmentation task on multi-channel images. Each channel represents a feature constructed from the order's location, satellite image, road networks, etc. E-partition \cite{guo2023towards} aims to cluster AOIs with equitable workload assignment. To achieve this goal, it first predicts the service time of a work given a specific AOI. Then, it clusters the AOIs into different delivery regions with the optimization goal based on the time prediction by setting the workload balance as the optimization goal.

\par Optimization-based models mostly resort to traditional optimization algorithms to achieve the service-semantic goals. Though promising, traditional optimization algorithms lack the ability to model different kinds of spatial and temporal features in the optimization process. These restrict their performance in real-life scenarios. This motivates us to propose a DRL-based framework that can handle different kinds of information as well as achieve the optimization goal. Moreover, the DRL-based model is more flexible, where different service-semantic goals can be introduced as rewards in the RL learning process. 





\section{Preliminaries and Problem Definition} \label{sec:problem_define}

\par In this section, we first give a general formulation for AOI segmentation under service-semantic goals. Then, we instantiate the general formulation by defining the AOI segmentation problem for the logistics service.

\subsection{A General Formulation }
\par  Without loss of generality, we first give a general form of service-semantic-aware AOI segmentation. It unifies the problem of previous efforts, where definitions and settings in previous methods can be viewed as different instantiations of the general formulation. 

\textbf{Definition 1: Geo-related Input.} The geo-related data $\mathcal{G}$ contains the graphical information of the target urban space, such as road networks or the satellite image of the target area. Incorporating geo-related data helps the model capture the geographical semantics.

\textbf{Definition 2: Service-related Input.} Massive historical data are generated in the service process. We use $\mathcal{S} = \{s_1, \dots, s_m\}$ to denote the service-related data, with each $s_i$ representing one type of data. For example, food orders in the online food delivery system or courier's trajectories in the logistics platform.

\textbf{Definition 3: Service-semantic Goals.} In different LBS services, there can be various service-semantic goals, such as predictability and workload equity, as introduced in Section~\ref{sec:related_work}. We use $\mathcal{O}=\{o_1, \dots, o_k\}$ to denote the service semantic goals, and each $o_i$ represents a predefined goal. 

\textbf{Definition 4: AOI Segmentation.} Given the geo-related data $\mathcal{G}$ and service-related data $\mathcal{S}$, service semantic-aware AOI segmentation learns a mapping function that can divide the target urban space into several AOIs, which aims to meet the service semantic goals $\mathcal{O}$, formulated as:
\begin{equation}
    {\mathcal{F}_{\mathcal{O}}({\mathcal{G}, \mathcal{S}}) \rightarrow A:=\{a_1, \dots, a_n}\},
\end{equation}
where $A$ is an segmentation result with $n$ AOIs, and $a_i$ means the $i$-th AOI.

\par In conclusion, the general formulation offers a shared understanding of the AOI segmentation problem, which recognizes the input and objective as geo/service-related information and service-semantic goals. This paves the way to create more effective models from the proposed perspectives.

\subsection{AOI Segmentation for Logistics Service}

\par Following the above general formulation, we instantiate the problem we focus on in this paper, i.e., AOI segmentation for logistics service, by specifying geo-related input, service-related input, and service semantic goals. 

\par The urban space is divided into several AOIs for efficient management of logistics services. Each AOI will be assigned to at least one courier, who will then be responsible for delivering all the packages within that AOI. Typically, the courier completes the delivery of all the packages in one AOI before moving on to the next one.

\textbf{Definition 5: AOI.}  To define the AOI, we first divide the geographic space into $M \times N$ grids, which efficiently and effectively represent the spatial unit, as adopted in many previous works. Each grid is represented by a tuple $(i,j)$, which means that the grid is in the $i$-th row and the $j$-th column. An AOI is defined as a collection of nearby grids $a=\{(i_1,j_1),\dots, (i_m,j_m)\}$, where $m$ is the number of grids contained in the AOI. 

\textbf{Definition 6: Road Network.} It is the geo-related input in our setting. We also use $\mathcal G$ to represent the road network to ease the presentation. It contains all roads within the region, formulated as $\{r_1,\dots,r_s\}$, where $s$ means the total number of roads. Each road $r$ is represented by a series of points,  i.e., $r=\{(x_{1},y_{1}),\dots,(x_{l},y_{l})\}$, where $x$ and $y$ means the longitude and latitude, respectively. $l$ is the number of points in the road $r$.


\textbf{Definition 7: Courier Trajectory.} It is the service-related input in our setting. A courier trajectory $\tau$ is a sequence of coordinates and time, i.e., $\tau=\{(x_1,y_1,t_1),...\}$, where $x$ is the longitude, $y$ is the latitude and $t$ is the time. We use $\mathcal{T}$ to represent the trajectories of all couriers. Section~\ref{sec:trajrl4aoi} details the motivation for introducing the trajectory.

\textbf{Definition 8: Delivery-service Semantic Goals.} We envision two service-related goals for the AOI segmentation in the package delivery service.

\par \textit{Goal 1: Trajectory modularity.} We borrow the concept of modularity from community detection in network analysis \cite{clauset2004finding}. A large modularity instructs dense connections within the same community but sparse connections between communities. Similarly, we want a large trajectory modularity to minimize switches between AOIs. Because in real situations, a courier will move to the next AOI after completing all the delivery tasks in one AOI. If the courier frequently switches between two nearby AOIs, then the two AOIs can be combined into one to form a more reasonable segmentation. The goal can be formulated as:
\begin{equation}
    o_1: \underset{A}{min} \underset{\tau}\sum{N_{switch}(A,\tau)},
\end{equation}
where ${N_{switch}(A,\tau)}$ is the number of switches between AOIs $A$ in the trajectory $\tau$.

\par \textit{Goal 2: Matchness with the road network.}
Road-network-based methods provide an initial segmentation (denoted by $A_{\mathcal{G}}$) that grasps the basic geographic-semantic meaning of the urban space. To this end, it is unreasonable if the result is quite different from road-based segmentation. We use the similarity between result and road-based segmentation to describe this semantic goal. It can be formulated as:
\begin{equation}
    o_2: \underset{A}{max} \, {\rm Similarity}(A,A_{\mathcal{G}}),
\end{equation}
where $\rm Similarity$ is a function that calculates the similarity between two segments, and $A, A_{\mathcal{G}}$ are the model result and segmentation based on the road network, respectively.

\textbf{Definition 9: AOI Segmentation Problem for Logistics Service.} Given the road network $\mathcal{G}$, and the courier's trajectory $\mathcal{T}$, we aim to learn an AOI segmentation that can satisfy the delivery-service semantic goals $\mathcal{O}=\{o_1, o_2\}$, formulated as:

\begin{equation}
    {\mathcal{F}_{\{o_1, o_2\}}({\mathcal{G}, \mathcal{T}}) \rightarrow A:=\{a_1, \dots, a_n}\}.
\end{equation}
\par To provide a big picture of our problem setting and the related works, we summarize their geo-related information, service-related information, and service-semantic goals in Table~\ref{tab:problem_setting_compare}.

\begin{table}[htbp]
	    \centering
	    \caption{Settings of different models under the general formulation.}
    	\resizebox{\linewidth}{!}{
    		\begin{tabular}{ccccc}
    			\toprule
    			Method & Geo-related Info  & Service-related Info & Service-semantic Goals \\
    			\midrule

                    \multirow{2}{*}{RegionGen \cite{Chen2023kdd}} &road networks & historical service  & predictability, \\
                    & & records & granularity, specificity\\
                    \midrule

                    \multirow{2}{*}{E-partition \cite{guo2023towards}} &road networks & order amount,  & equitable workload  \\
                    & & service time & assignment\\
                    \midrule

                    \multirow{2}{*}{C-AOI \cite{zhu2023c}} &road networks, & order  &  closer to   \\
                    & satellite images &  locations & geographic boundaries \\ 

    			\midrule
    			\midrule

                    \multirow{2}{*}{Our Model} &road networks & courier   & trajectory modularity,  \\
                    & & trajectory & Matchness with road networks\\
    			\bottomrule
    		\end{tabular}
    	}
	\label{tab:problem_setting_compare}
\end{table}



\section{Proposed DRL4AOI Framework}
\par This section describes the DRL4AOI framework, a deep reinforcement learning (DRL) framework for AOI segmentation. We first introduce how we formulate the problem from the RL perspective and then present the details of the framework.

\subsection{Formulation from the RL perspective}
\par
\begin{figure}[htbp]
    \centering
    \includegraphics[width=0.9 \linewidth]{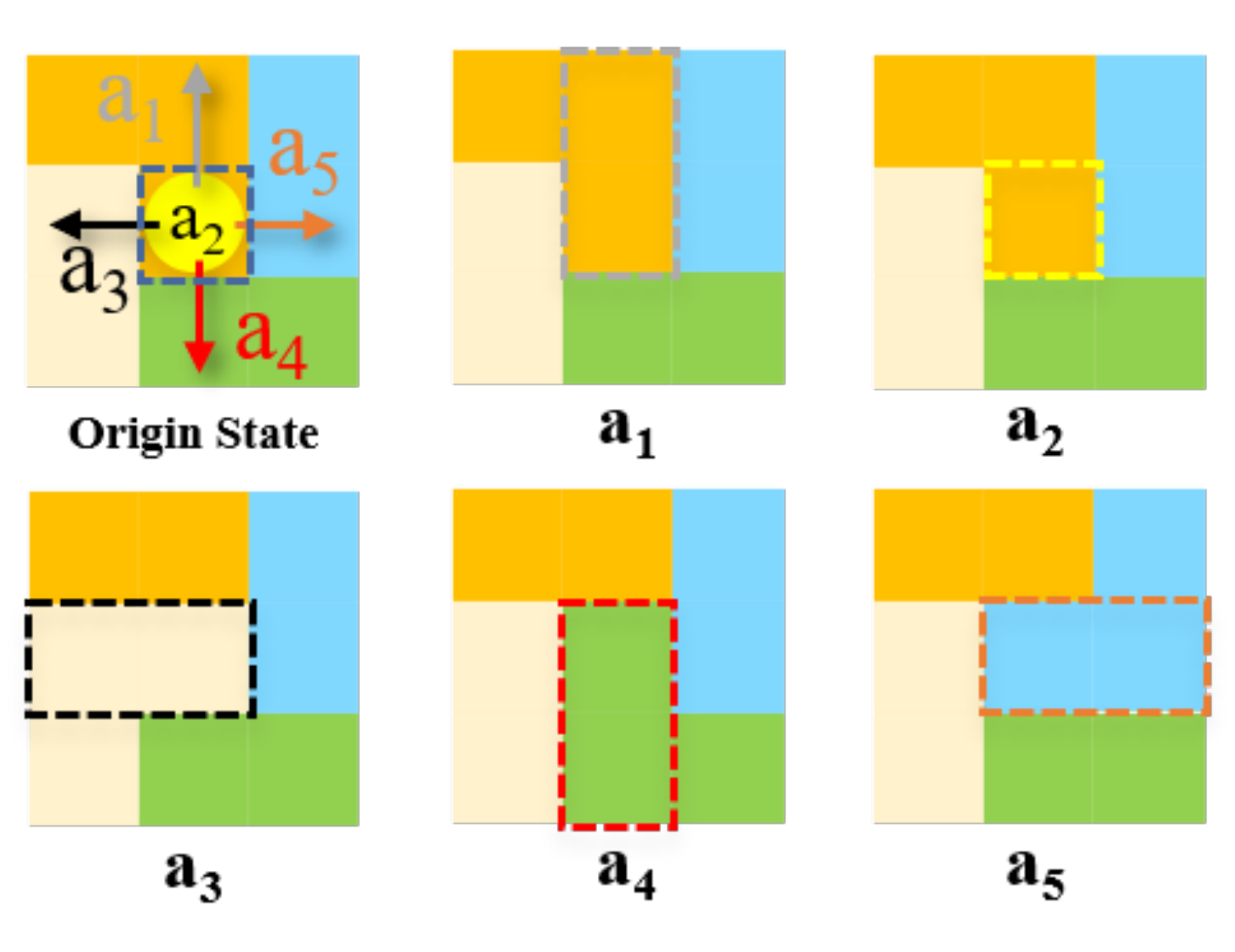}
    \caption{Illustration of Action. It is defined as merging the target grid to its up/down/left/right AOI or staying in its current AOI.}
    \label{fig:action}
\end{figure}

 We provide a novel perspective to show that the problem can be naturally regarded as a sequential decision-making problem, which can be effectively mitigated by reinforcement learning. As illustrated in Figure~\ref{fig:action}, the core idea behind this is deciding which neighbor AOI it belongs to for each grid on AOI's border at each decision step (To ease the presentation, we call this action AOI selection).

\par Specifically, the process of making sequential decisions can be represented by a finite-horizon discounted Markov Decision Process (MDP) \cite{sutton1998introduction}. In this process, an AOI segmentation agent interacts with the environment over $T$ discrete time steps by the action of AOI selection. Formally, a MDP can be formulated as $M = (\emph{S}, \emph{A}, \emph{P}, \emph{R}, \gamma)$, where $\emph{S}$ is the set of states, $\emph{A}$ is the set of actions, $\emph{P}: \emph{S} \times \emph{A} \times \emph{S} \rightarrow {\mathbb R}_{+}$ represents the transition probability, and $\emph{R}: \emph{S} \times \emph{A} \rightarrow {\mathbb R}$ represents the reward function. The initial state distribution is $s_0: \emph{S} \rightarrow {\mathbb R}_{+}$, and $\gamma$ is a discount factor whose value is between 0 and 1. Moreover, it is worth mentioning that by setting the action as adjusting the AOI's border, the spatial continuity constraint that each AOI should have inter-connected grids can be naturally and easily satisfied.

When provided with a state $s_t$ at a given time $t$, the AOI segmentation agent utilizes the current policy $\pi_\theta$ which is a deep neural network parameterized by $\theta$ to generate an action, i.e., selecting a nearby AOI for the grid on AOI's border. The agent then receives a reward $r_t$ defined by the service semantic goals from the environment. Then, the reward is utilized to train the agent. During the training process, the main objective is learning the best parameter $\theta^{*}$ for the agent to maximize the expected cumulative reward, which is formulated as:
\begin{equation}
    \theta^{*} = {\arg\max}_{\theta}{\mathbb{E}_{\pi_{\theta}}\left[\mathop\sum\limits_{t=1}^{T}{\gamma^t}r_t\right]},
\end{equation}
where the discount factor $\gamma$ controls the trade-offs between the importance of immediate and future rewards, and the total time step, denoted by $T$, is determined by the number of grids in AOI's border. The following part introduces the details of the agent, state, action, reward, and state transition probability. 

\par \noindent \textbf{AOI Segmentation Agent}: The AOI segmentation agent is responsible for learning the target function $\mathcal{F}_{\mathcal O}$. It accomplishes this by selecting an AOI for each grid on the AOI border with the help of a deep neural network, which will be detailed in the action introduction part. The agent follows an encoder-decoder architecture, where the encoder takes meaningful features from the current state $s$ and embeds them into hidden representations. The decoder accepts these representations as input and produces an action $a$ at each time step. Abstractly, the agent can be represented as:
\begin{equation}
    a_t = {\rm Decoder}({\rm Encoder}(s_t)),
    \label{eq:framework_agent_encoder}
\end{equation}
where ${\rm Encoder}$ and ${\rm Decoder}$ represent the state encoder and action decoder of the agent, respectively. 

\par \noindent \textbf{State}: A state $s_t$ is composed of the information of current AOI segmentation and the target grid in the AOI border to be modified.  All the possible segmentations make up the entire state space. 

\par \noindent \textbf{AOI-select Action}: An action  $a_t \in \mathcal{A}_t$ is defined as selecting a nearby AOI for a grid on the AOI border. Intuitively, we can set all current AOIs as candidates in the action space, i.e., $\mathcal{A}_t = \{a_1, \dots, a_n\}$. However, such a design introduces a large search space, which challenges the agent to learn a converged policy. To address the issue, we consider the AOI selection as merging the target grid into its neighbor AOIs that are located in different directions, as shown in Figure~\ref{fig:action}. In this way, the action space is reduced to only five actions, namely, up, down, left, right, and origin. Where up/down/left/right means merging the target grid to its up/down/left/right AOI, and origin means that the target AOI stays in its current AOI. We use a five-dimensional one-hot vector to represent an action in the implementation.

\begin{figure*}[!t]
      \centering
      \includegraphics[width=1\linewidth]{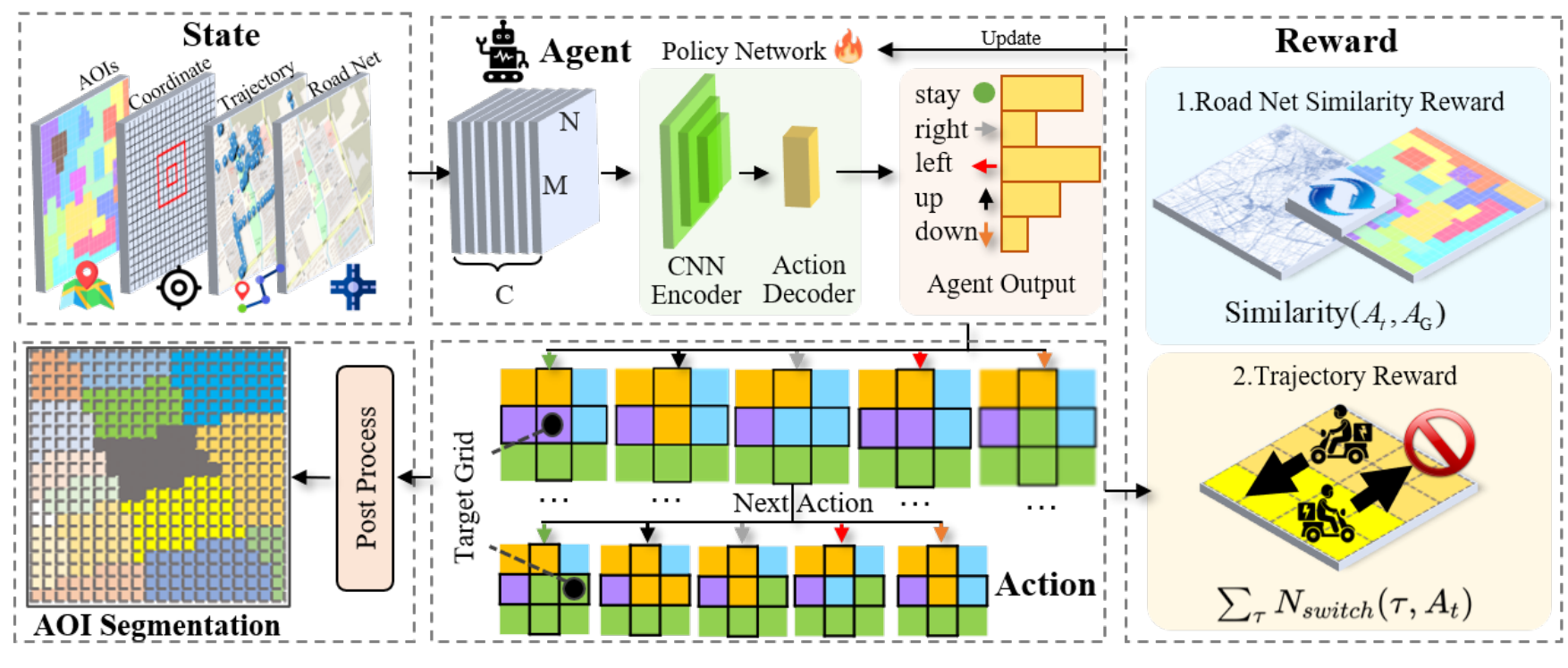}
      \caption{DRL4AOI Architecture, which mainly includes three steps: 1) Data preprocessing, which rasterizes the map and prepares service-related input. 2) RL-based AOI segmentation, which gradually adjusts the grids in AOI's border by the agent under the guidance of the rewards designed by service-semantic goals. 3) Post-processing, which is designed to make further refinement to the segmentation results.}
      \label{fig:model}
      \vspace{-1.5em}
\end{figure*}

\par \noindent \textbf{Service-semantic Reward}: \( r_t \in \mathbb{R} \leftarrow \emph{S} \times \emph{A} \): Reward design is of great importance in reinforcement learning; a well-designed reward can have a significant impact on the learning performance of the agent and ultimately determine the quality of the model. Recall that in the AOI segmentation problem, some service-specific semantic goals need to be achieved after the partition. Those goals can be naturally converted into rewards in the RL framework. This mechanism introduces the flexibility to accommodate various goals in different services. In Section~\ref{sec:trajrl4aoi}, we will introduce how to design the reward for logistics service.

\par \noindent \textbf{State Transition Probability}:  $P(s_{t+1}|s_t, a_t): \emph{S} \times \emph{A} \times \emph{S} \rightarrow {\mathbb R}_{+}$ represents the transition probability from state $s_t$ to $s_{t+1}$ if action $a_t$ is taken at $s_t$, which means that the state transfer from one segmentation from another one. In the AOI segmentation problem, the environment is deterministic with no uncertainty in the state transition, which means the state $s_{t+1}$ transited from state $s_t$ after taking action $a_t$ is deterministic.

\subsection{DRL4AOI Framework Architecutre}
\par Based on the above formulation, we are now ready to introduce the proposed framework, DRL4AOI. Figure \ref{fig:model} shows the overall architecture of DRL4AOI, where the inputs are geo-related and service-semantic-related information, and the output is the result of AOI segmentation. The RL-based segmentation learning mainly includes three steps: 1) data preprocessing, 2) model training, and 3) post-processing. 

\par Firstly, the data preprocessing step constructs both the geo-related input and service-related input for the model. Secondly, the RL-based AOI segmentation solves the problem from the reinforcement learning perspective. Specifically, this step trains an AOI segmentation agent, which takes the state as input and outputs an AOI selection  (up/down/left/right/origin) at each time step. After each action is performed, the environment will provide reward feedback that is aligned with service-semantic goals. Guided by the rewards, the agent updates its parameters and gradually adjusts the grids in AOI's border guided. In order to avoid the influence of traverse order on the partition result, the agent will traverse several times in grids. As the training progresses, the accumulated reward will gradually rise, which means a better segmentation strategy is learned regarding the service-semantic goals. At last, the post-processing further refines the segmentation results outputted in the previous step. We give a detailed introduction in the following part.

\section{Proposed TrajRL4AOI For Logistics Service} \label{sec:trajrl4aoi}


\par Based on the DRL4AOI framework, we further propose a model called TrajRL4AOI for logistics delivery service to demonstrate the effectiveness and generality of the framework. TrajRL4AOI incorporates two types of input - the road networks as the geo-related information and the courier's trajectory as the service-semantic input. Since they are the most common data in the realm of spatial-temporal data mining \cite{hu2019stochastic, yang2013using, yuan2014discovering,yao2022trajgat}. TrajRL4AOI uses these to generate high-quality AOIs by setting the trajectory modularity and road network matchness as the service-semantic goals.


\subsection{Motivation of Introducing the Trajectory}

\par
\begin{figure}[htbp]
    \centering
    \includegraphics[width=1 \linewidth]{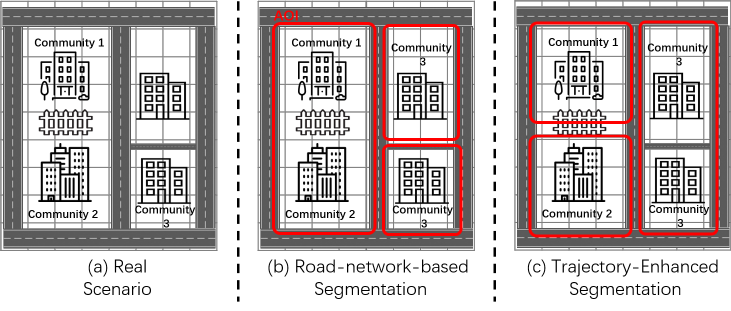}
    \caption{Motivation of introducing trajectory. Two communities (i.e., Com1 and Com2) are surrounded by the same road network, separated by a fence. It is unreasonable to divide them into the same AOI according to the road network. By leveraging a vast amount of courier trajectory data, we can better distinguish between Com1  and Com2, resulting in more accurate and suitable AOI segmentations.}
    \label{fig:traj_motivation}
\end{figure}

  Before introducing TrajRL4AOI, it is worth introducing the motivation for modeling the courier's trajectory. As shown in Figure~\ref{fig:traj_motivation}(a), two communities (i.e., Com1 and Com2) are surrounded by the same road network, while a fence separates them. And there is another big community (i.e., Com3) which has a small road inside. If we generate the AOI segmentation by road-network-based model (the result is shown in Figure~\ref{fig:traj_motivation}(b)), it can be seen that Com1 and Com2 are assigned to the same AOI. However, it is better to identify them as two separate AOIs in the logistics system since it is more aligned with the real scenario. Similarly, we also hope that all components of Com3 can be merged into the same AOI while road-network-based models fail to do so. To mitigate the problem, we leverage the trajectory to refine the AOI segmentation. As shown in Figure~\ref{fig:traj_motivation}(c), if we take couriers' trajectories into account, it can be seen that the courier's trajectory would never switch across the fence between two communities. In that case, the two communities can be easily detected as two AOIs. In the same way, we can merge the two parts of Com3 into one AOI by the observation that the courier often switches between the two parts. In summary, introducing trajectories can improve the AOI segmentation alignment with the real service process. In the next part, we present the details of each step in TrajRL4AOI and how the courier's trajectory is incorporated into the RL framework.

\subsection{Preprocessing} \label{sec:preprocessing}
\par The data preprocessing contains three kinds of feature processing, including AOI, trajectory, and road network.
\par \noindent \textbf{AOI features processing.} We rasterize the urban space to represent AOI segmentation. Specifically, we rasterize the map into ${M\times N}$ grids to facilitate the model analysis and processing, where $M, N$ is the length and width after rasterization. In our problem, using a grid to represent the smallest unit in the map offers the following two merits:  1) Grid mapping is an effective way that naturally ensures coverage of all location points within a city; 2) The AOI adjustment can be straightforwardly achieved through grid aggregation or splitting, which can get more fine-grained optimization especially compared to road-network based models. Because in road network-based models, the smallest operational units are geographical objects that are divided by the road network. Some of the geographical objects may deserve further segmentation.

\par \noindent \textbf{Trajectory processing.} To represent and mine the mobility patterns of couriers, we further build a trajectory transfer graph $\mathcal{T}_{G}$ based on all couriers' trajectory data. It is a directed weight graph that describes the trajectory transition between nearby grids.  $\mathcal{T}_{G}$ can be formulated as $\mathcal{T}_{G}=(V, E)$, where $V$ is the node set with each node corresponding to a grid in the gridded map. And $E$ is the edge set, with each edge represented by a triplet $(s,d,w)$, which means there are total $w$ trajectories from $s$-th node to $d$-th node. We flatten the grids to obtain the 1D index of 2D grids. In order to get
 $E$, we map each point in every trajectory $\tau$ into a grid and sort them in time order. So that it is converted into $\tau=\{n_1,\dots,n_m\}$, where $n$ means the point is mapped into the $n$-th grid, and $m$ is the total point number in trajectory $\tau$.




\par
\begin{figure}[htbp]
    \centering
    \includegraphics[width=1 \linewidth]{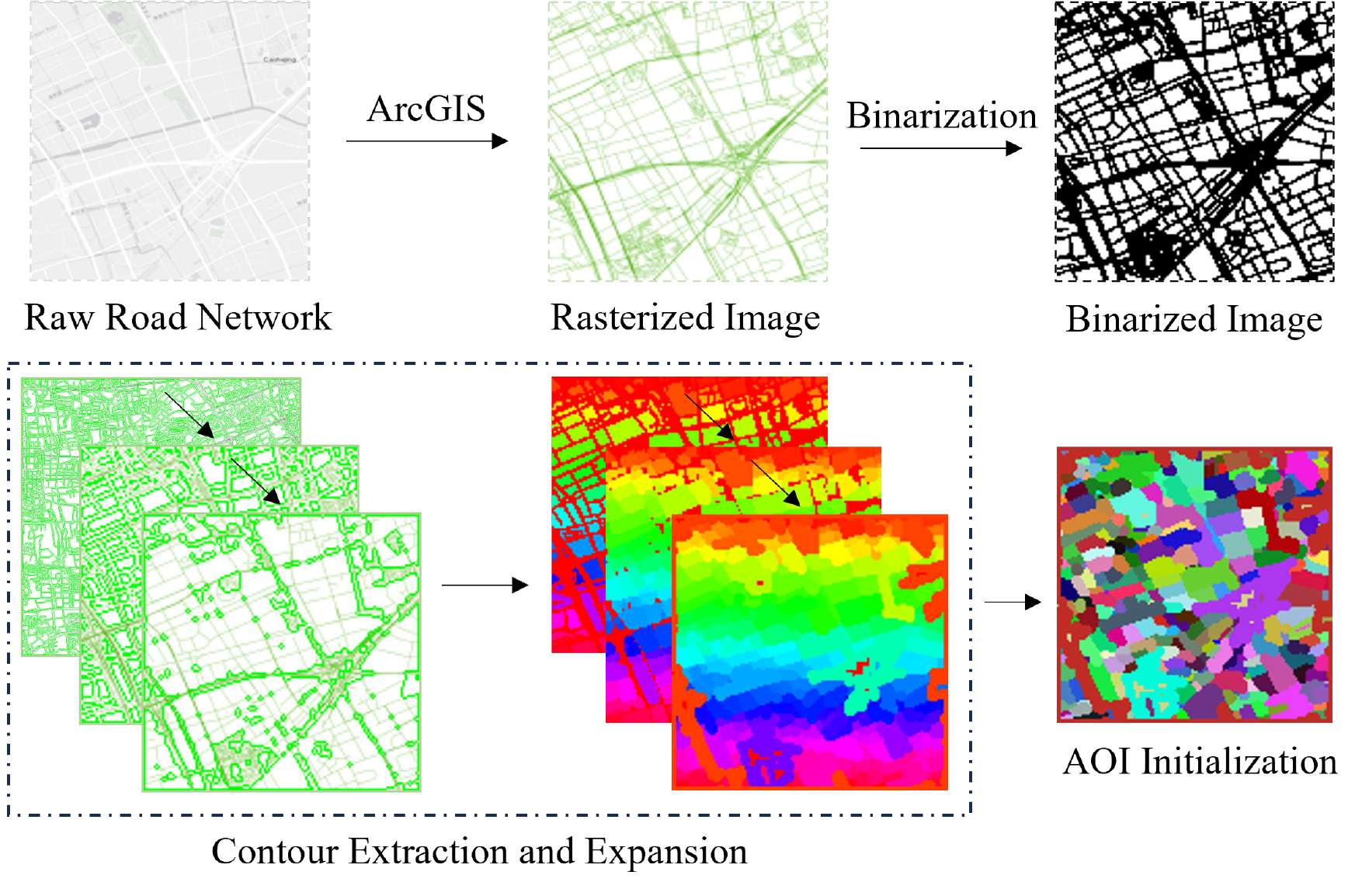}
    \vspace{-1em}
    \caption{AOI initialization based on the road network. It contains two steps: 1) obtain the road network data and export it as an image. 2) perform binary thresholding, connected components, and contour extraction and expansion.}
    \label{fig:road network aoi init}
    \vspace{-1em}
\end{figure}

 \noindent \textbf{Road Network processing.} Figure \ref{fig:road network aoi init} shows the process of obtaining road network data and converting it into an initial AOI segmentation. It can be divided into two steps:

Step 1: Rasterizing the road network data and exporting the road network pixel matrix. We obtain the required road network data from OpenStreetMap\footnote{https://www.openstreetmap.org/} \cite{xu2023road} and import it into the ArcGIS\footnote{https://www.arcgis.com/} software. In ArcGIS, we crop the road network information based on the desired coordinate range, remove all other elements, and only keep the road network. Then, we export it as an image format with a certain precision for further processing.

Step 2: Performing semantic segmentation on the road network image to obtain initial AOI segmentation. We convert the map image to grayscale and perform binary thresholding to distinguish the road network from other elements clearly. The road parts in the map are marked as black, and the rest are white. Then, we find connected components\footnote{https://docs.opencv.org/} on the binary image and obtain a label matrix containing AOI regions separated by the road network. To remove the road network itself from the AOI segmentation, we perform contour extraction and expansion \cite{Chen2023kdd} on the non-road regions iteratively. In each iteration, we extract contours and expand the label matrix based on the contour information. Pixels within the same contour are assigned the same label. Through multiple iterations, neighboring regions with similar features are merged into the same region.

By following these steps, we obtain the initial AOI segmentation of the road network, which can be used as input to the model.

\subsection{Double-DQN for AOI Segmentation}

\par In this subsection, we elaborate on the proposed TrajRL4AOI by introducing three important components in the framework, including the state, reward, and agent. 

\par \noindent \textbf{State}. A state records information around the target grid, the current AOI segmentation, trajectory and road network partition. Here we use a matrix with size $M\times N$ to describe all grids in the map. Generally, a state $s \in \mathbb{Z} ^{M \times N \times C}$ is a feature matrix with $C$ channels. The following features are extracted to achieve a comprehensive representation of the current state: 

\begin{itemize}
     \item Target grid, denoted by $x_t \in \mathbb{B}^{M \times N \times 1}$, which records the target grid (i.e., the grid to be modified) in the receptive field by a $M \times N$ boolean matrix. The position of the grid that needs to be modified is set to 1, while the other positions are set to 0.
     
    \item AOI features $A_t \in \mathbb{Z}^{M \times N \times 1}$. It records the current AOI segmentation (i.e.,  AOI ID) of all grids in the $M \times N$ field. Understanding its neighbor AOI of a grid is crucial for the decision of AOI selection.
    
    \item Trajectory features ${\mathcal{M}}_{\mathcal{T}} \in \mathbb{Z}^{M \times N \times 4}$, which records the local trajectory transfer in the $M \times N \times 4$ field. $a_{i,j,d}$ means the trajectory number of the grid in $i$-th row, $j$-th column and the direction $d$ (i.e., up/down/left/right). Such information illustrates how grids interact with their neighbors regarding the trajectory connection in the courier's service process \cite{wang2021trajectory}.

    \item Road segmentation features ${\mathcal{G}}_{t} \in \mathbb{Z}^{M \times N \times 1}$. It contains the AOI segmentation results based on the road network in the $M \times N$ field. The information on road networks could help the AOI selection avoid some abnormal divisions, such as narrow areas.
\end{itemize}
\par At last, the overall state at time step $t$ is represented by the concatenation of the above features, formulated as  $s_t = (x_t, A_t, \mathcal{M}_{\mathcal{T}}, \mathcal{G}_t)$.

\par \noindent \textbf{Action\&State Transition.} As defined in the framework, actions indicate the nearby AOI to which the target grid should belong, which is transformed from the direction indicator (up, down, left, right, origin.) in the action space. Once an AOI selection is made for the target grid, the current AOI segmentation will change, thus forming a state transition.



\par \noindent \textbf{Reward Design}. One advantage introduced by the reinforcement framework is the flexibility of the reward design, which can accommodate different service-semantic goals. In the logistics service, the two goals are trajectory modularity and road network matchness, as we introduced in Section~\ref{sec:problem_define}. In that case, the reward $r_t$ after an action is taken at step $t$ is composed of two parts in TrajRL4AOI, with each $r_t^i$ defined as follows:

\begin{itemize}
    \item Trajectory reward $r_t^1$. This reward is related to the objective $o_1$ (i.e., trajectory modularity). Recall that ${N_{switch}(\tau, A)}$ is the number AOI switch for trajectory $\tau$ under current AOI segmentation $A$. Let $o_1(s_t) = \underset{\tau}\sum{N_{switch}(\tau, A_t)}$  be the total switch times of all trajectories at state $s_t$. The trajectory reward is defined as the difference between the objective values $o_1(s_t)$ and $o_1(s_{t+1})$, i.e., 
    \begin{equation}
        r_t^1 = o_1(s_t) - o_1(s_{t-1}).
    \end{equation}

    \item Similarity to road-based segmentation $r_t^2$, which is related to the second objective $o_2$, i.e., Appropriate segmentation based on the road network. Let $o_2(s_t)=Similarity(A_t,A_{\mathcal{G}})$, where $A_{\mathcal{G}}$ denotes road network based division. And $Similairy$ calculates the Co-AOI rate (detailed in Equation \ref{eq:CR}), i.e., the similarity between current AOI segmentation and segmentation based on the road network. The reward $r_t^2$ is defined as the similarity change between two states, i.e.,
    \begin{equation}
        r_t^2 = o_2(s_t)-o_2(s_{t-1}).
    \end{equation}

 
\end{itemize}

\par Overall, the total reward function is defined as:
\begin{equation}
    r_t = k_1 * r_t^1 + k_2 * r_t^2
\end{equation}
where $k_1, k_2 >0$ are hyper-parameters to control the weight of different rewards.

\par \noindent \textbf{Agent}. The agent is designed to learn a policy network for AOI selection. We adopt Double Deep Q-Networks (DDQNs) \cite{van2016ddqn} to implement the agent because of its strong power of combining deep neural networks with reinforcement learning, as proved in different environments such as game playing \cite{mousavi2018deep, mnih2013dqn, goldwaser2020deep}. As shown in Figure~\ref{fig:model}, at each step, the agent encodes the state with a Convolutional Neural Network (CNN) \cite{o2015cnnintro, li2021cnn}. Via multiple convolutional operations, the CNN state encoder extracts an effective representation of the target grid's neighborhood information, which will be further fed into the policy networks.  The AOI-selection policy network takes the state embedding as input and scores each action candidate with a Multi-Layer Perceptron (MLP)\cite{taud2018mlp}, where the obtained score indicates the performance score of the action in the learned policy.

Specifically, DDQN is based on DQN \cite{mnih2013dqn, mnih2015dqn}, whose primary idea is to approximate the Q-value function using a neural network. The network takes a state \( s \) as input and produces Q-values for all possible actions. The key equation for the Q-value in DQNs is given by the Bellman equation:
\begin{equation}
    Q(s, a) = r + \gamma \max_{a'} Q(s', a')
\end{equation}
The optimal action value Q is the sum of the reward and the maximum Q-values of all actions in the next state.

However, DQNs tend to overestimate Q-values because they use the same network to select the best action and evaluate it. This can lead to suboptimal policies and reduced stability in training.

Double DQNs aim to address this overestimation bias. Instead of using a single network for both action selection and evaluation, Double DQNs use two networks: i) An online network to select the best action. ii) A target network to evaluate the Q-value of the selected action. Thus, the updated Q-value equation in Double DQNs is:
\begin{equation}
    Q(s, a) = r + \gamma Q_{\text{target}}(s', \text{argmax}_{a'} Q_{\text{online}}(s', a'))
\end{equation}

Here, \( Q_{\text{online}} \) is the Q-value estimated by the online network, and \( Q_{\text{target}} \) is the Q-value estimated by the target network. The optimal action value Q is determined by the two components: the immediate reward, and the Q-value from the target network of the best action in the next state. The best action is the action of the maximum Q-value in the online network. By decoupling the action selection from its evaluation, Double DQNs reduce the overestimation bias, resulting in more stable and efficient learning.

Furthermore, DDQN retains all other innovations from DQN, including experience replay and the usage of a target network for stabilizing updates. Through the decoupling of action selection and Q-value evaluation, DDQN offers a more conservative estimate of Q-values, which in turn provides a notable improvement in performance and stability over vanilla DQN in various tasks. The post-processing further refines the segmentation result to avoid the fragmented partition. We merge each AOI to one node in the result and calculate the new trajectory transfer graph. Then, the Louvain\cite{blondel2008fast} algorithm is used for post-processing.


\subsection{Discussion}
\subsubsection{Addtional Geo-/Service-related data}
In reality, additional data may be available which can contribute to AOI segmentation. For example, geo-related data such as satellite image \cite{bins1996satellite, dare2005satellite}, POI distribution \cite{ye2011poi, liu2013poi}. In terms of the service-semantic data, different data can exist in different service scenarios, such as the courier's trajectories \cite{ruan2022couriers, dodge2012couriers}, and the order information \cite{liu2021order} in the food delivery. Considering this, we improve the flexibility of DRL4AOI from the following two aspects: i) RL combined with deep reinforcement learning. The deep neural network-based nature allows the model to accommodate abundant information from different domains. For example, the satellite image can seamlessly serve as a feature channel in the input of the CNN. ii) customzied  AOI segmentation agent encoder.  In our work, the input is organized into grid data. Thus, the CNN-based encoder is utilized to represent the information in the input. To allow for more types of data structure, the AOI segmentation agent encoder can be customized, such as adding the graph neural network for graph-based data. 

\subsubsection{Addtional service-semantic goals}
\par Moreover, various service-semantic goals can exist in real scenarios, mainly due to various LBS service types and the development of the service.  Those service-semantic goals can be broadly classified into two classes: the first one directly relates to AOIs themself, which requires specified attributes the AOI needs to have, such as proper granularity  \cite{Chen2023kdd}, or matchness with geographic boundaries \cite{zhu2023c}. The other class is related to the service performance based on the AOI segmentation results, such as predictability \cite{Chen2023kdd} and workload equity \cite{guo2023towards}. In our designed framework, different goals can be considered in a plug-and-play way by converting them into different rewards. By weighing the importance of different rewards, the segmentation agent tunes its policy to set its preference for different goals. In summary, the reward mechanisms serve as a convenient way to bridge the service-semantic goals and the model training.

\section{Experiment}


\subsection{Experiment Setting}


\par \noindent \textbf{Data.} We conduct extensive experiments on synthetic data:

\par
\begin{figure}[htbp]
    \centering
    \includegraphics[width=1 \linewidth]{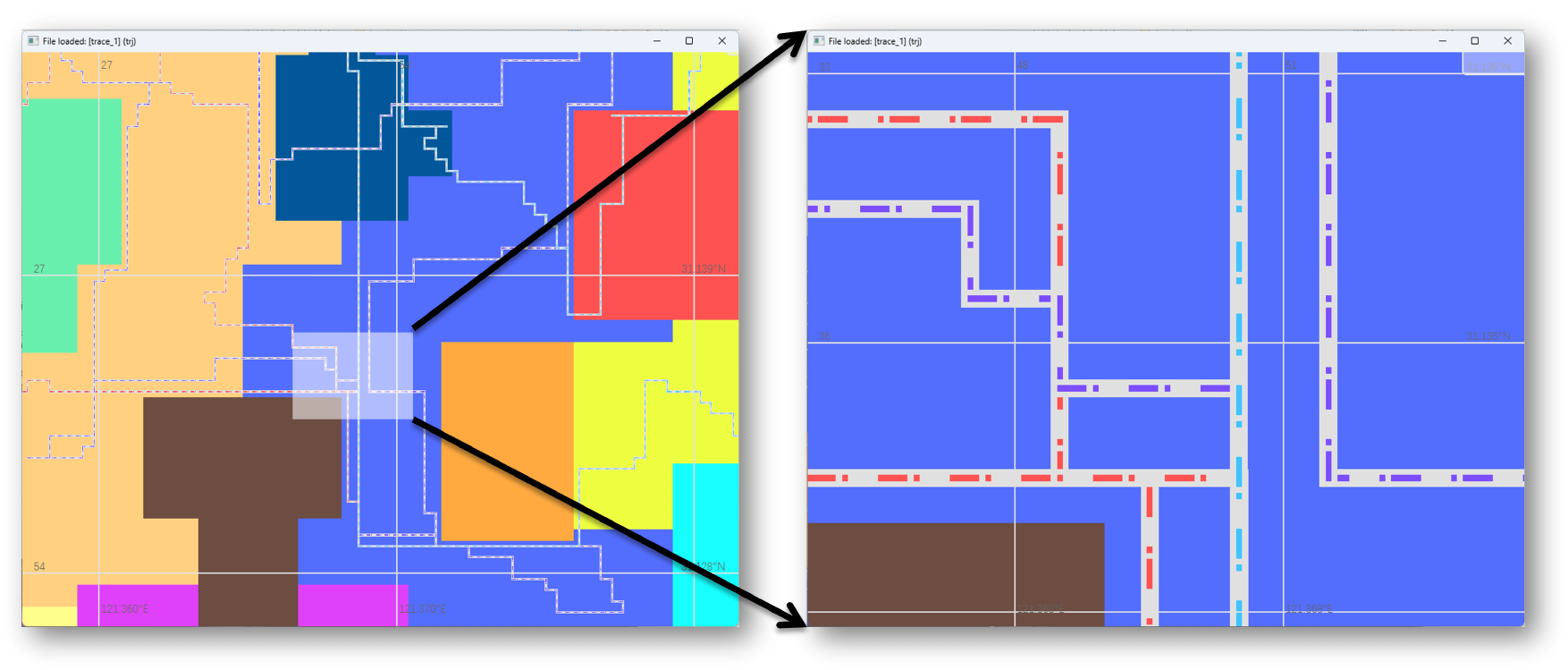}
    \vspace{-1em}
    \caption{Illustration of Generated AOI and trajectories. It is a 100 $\times$ 100 grid that contains 20 AOIs, with trajectories both inside an AOI and traveling between different AOIs.}
    \label{fig:aoi_example}
\end{figure}

  \textit{Synthetic Data.} To test the performance of different methods in the ideal case, we first create a benchmark synthetic dataset called SyntheticAOI. The main idea is to generate ground-truth AOIs and then trajectories under the AOIs. The data construction mainly contains two steps: 1) AOI generation. We first generate several AOIs for $M\times N$ grids and assign them to several couriers. Each courier is responsible for one AOI. 2) Trajectory generation. We randomly generate several trajectories for each courier based on the previously generated AOI. To generate a single trajectory, we first randomly generated several packages in his served AOIs, and generated the trajectory by solving a VRP problem \cite{bello2016neural, braekers2016vrp} with the minimized total distance. We give an example in Figure~\ref{fig:aoi_example}.  

\vspace{-1em}

\par \textit{Real-world Data.} Our research is based on the delivery data provided by one of the world's largest logistics companies, which includes data from Hangzhou and Shanghai, China. The dataset comprises both order information and trajectory data. We collected data from Jun. 2019 to Aug. 2019, resulting in nearly 260,000 orders and 57.04 million trajectory points. The trajectory data was sampled at intervals of 2-4 seconds. For our study, we extracted the address latitude, longitude (GCJ-02 coordinate system), and courier ID from the order and trajectory data.

\par \noindent \textbf{Metrics.} For synthetic data, we have the ground-truth AOI segmentation. Therefore, one intuitive way of evaluation is calculating the similarity between the ground-truth segmentation and segmentation given by the model. Following this idea, we introduce the following metrics to evaluate the performance of different models in synthetic data:

\begin{itemize}
     \item Fowlkes-Mallows Score (FMI). It calculates the similarity between two partitions. It is the geometric mean of every grid pair's precision and recall rates in two partitions, which is formulated as:
    \begin{equation}
        FMI = \frac{TP}{\sqrt{(TP+FP)} \sqrt{(TP+FN)}},
        \label{eq:FMI}
    \end{equation}
    Where $TP, FP$ and $FN$ are metrics in the confusion matrix. Specifically, given any two grids $g_1=(i_1, j_1)$ and $g_2=(i_2, j_2)$, it is said to be equivalent if $g_1$ and $g_2$ belongs to the same AOI. Otherwise, it is said to be non-equivalent. The $TP$ is the total number of grid pairs, which are equivalent both in the model's output $\hat A$ and the ground truth AOI $A$. The $FP$ is the number of grid pairs that are equivalent in $\hat A$ but non-equivalent in $A$. And the $FN$ is the number of grid pairs that are both non-equivalent in $A$ and $\hat A$.
    
    We referred to the evaluation metrics of clustering algorithms because this problem can also be regarded as a clustering problem of grids. Note that since the vector representation of each grid point cannot be obtained, internal evaluation metrics cannot be used to indicate the quality of the partition.
    
    \item Co-AOI Rate (CR). It calculates the similarity between two AOI segmentations from the grid-pair perspective. Specifically, for any grid pairs, $g_1=(i_1, j_1)$ and $g_2=(i_2, j_2)$, it is said to be concordant if $g_1$ and $g_2$ are equivalent both in $A$ and $\hat A$. Otherwise, it is said to be a discordant pair. When $(g_1, g_2)$ is a concordant pair, we can also say that they are Co-located in the same AOI both in prediction and ground truth. Therefore, the Co-AOI Rate calculates the ratio of concordant pairs in all pairs, which can be formulated as:
    \begin{equation}
        {\rm CR} = \frac{ \underset{g_i,g_j \in A}{\sum} {\mathbb I}_{A,\hat{A}}(g_i, g_j)}
        { \underset{g_i,g_j \in A}{\sum} {\mathbb I}_{A}(g_i, g_j)},
        \label{eq:CR}
    \end{equation}
    where ${\mathbb I}(\cdot)$ is the indicator function, and $\mathbb I_{A,\hat{A}}({\hat \pi}_i, \pi_i)$ equals 1 if $g_i$ and $g_j$ are co-located in the same AOI both in the segmentation $A$ and $\hat{A}$, else 0. Similarly, $\mathbb I_{A}({\hat \pi}_i, \pi_i)$ equals 1 if $g_i$ and $g_j$ are co-located in the same AOI in segmentation $A$.
\end{itemize}
\par Besides, we also choose the performance in the two semantic goals as the metrics, named $R_1$, which evaluates the trajectory switch between AOIs, and $R_2$, which measures the similarity between road network segmentation. Since $R_2$ uses the result of road network segmentation as the criterion, we only report the $R_2$ of other models. Model performance is better if these metrics are higher.

\begin{table*}[!t]
    \center
    \caption{Results in Synthetic Dataset.}
    \begin{threeparttable}
    \begin{tabular}{*{13}{c}}
        \toprule
        & \multicolumn{4}{c}{$5 \times 5$} & \multicolumn{4}{c}{$6 \times 6$} &
        \multicolumn{4}{c}{$10 \times 10$} \\ 
        \cmidrule(lr){2-5} \cmidrule(lr){6-9} \cmidrule(lr){10-13}
       ~ & $R_1$ & $R_2$ & FMI & CR & $R_1$ & $R_2$ & FMI & CR & $R_1$ & $R_2$ & FMI & CR \\ \midrule
        GreedySeg & -15.1 & 36.7\% & 49.7\% & 36.7\% & -10.2 & 27.7\% & 58.0\% & 39.2\% & -13.8 & 24.9\% & 53.5\% & 31.4\% \\
        Louvain & -11.0 & 41.8\% & 56.0\% & 44.4\% & -11.3 & 26.4\% & 54.0\% & 35.0\% & -14.2 & 25.1\% & 52.8\% & 30.7\% \\
        DBSCAN & -13.1 & 47.0\% & 46.8\% & 48.9\% & -7.4 & \textbf{80.4\%} & 42.5\% & 59.9\% & -23.1 & 63.3\% & 39.2\% & 61.8\% \\
        CKMeans & -5.6 & \underline{54.0\%} & 45.8\% & 50.2\% & -9.4 & 48.2\% & 44.8\% & 49.3\% & -11.4 & \underline{67.7}\% & 58.4\% & 59.7\% \\
        GCLP & -8.0 & 29.1\% & 55.9\% & 36.7\% & -15.3 & 24.1\% & 45.4\% & 27.2\% & -24.0 & 16.0\% & 38.2\% & 17.9\% \\
        RoadNetwork & \underline{-4.0} & - & \underline{70.8}\% & \underline{73.2\%} & \underline{-5.6} & - & \underline{66.3}\% & \underline{69.6\%} & \underline{-4.9} & - & \underline{74.9}\% & \underline{75.9\%} \\
        \midrule
        TrajRL4AOI (Ours) & \textbf{-2.0} & \textbf{57.8\%} & \textbf{86.6\%} & \textbf{77.0\%} & \textbf{0.0} & \underline{68.6\%} & \textbf{100.0\%} & \textbf{100.0\%} & \textbf{-1.1} & \textbf{81.7\%} & \textbf{94.0\%} & \textbf{95.1\%} \\
        \bottomrule
    \end{tabular}
    \end{threeparttable} 
    \label{tab:result}
    \vspace{-1em}
\end{table*}

\par \textbf{Implementation.} We conduct the experiments and baselines on a machine with a Hygon C86 7151 CPU and NVIDIA RTX A4000 GPUs. We use the RMSprop optimizer for the parameter update. The initial learning rate is set to $10^{-4}$ and decays in each episode. The random seed is set to $3047$. The training episode is set to $500$. The DoubleDQN updates every 1000 steps on synthetic data and 10000 steps on real-world data. This frequency difference is due to the larger scale of real data, requiring more samples for training. We extract the features by applying $2$ layers of $5\times 5$ kernel. The hyper-parameters $k_1,k_2$ in the total reward function are set to $0.6, 0.4$, respectively.

\subsection{Baselines.}

We choose several methods that belong to different types as baselines:

\begin{itemize}
    \item GreedySeg. A simple method to satisfy the trajectory semantic goal. We traverse each grid and merge it into one nearby AOI, which can reduce the trajectory switch to the maximum extent.
    
    \item RoadNetwork \cite{lei2012districting}. A basic model that conducts the AOI segmentation based on the road network, as mentioned in Section~\ref{sec:preprocessing}. 

    \item DBSCAN \cite{ester1996density}. It calculates the AOI segmentation based on package clustering. Specifically, it divides all packages into several clusters, each representing one AOI. In this way, the AOI of a certain grid is assigned to the most nearby package center.

    \item CKMeans \cite{bradley2000constrained}. A clustering algorithm that allows a given range of cluster numbers. Firstly, CKMeans clustering is applied to all the packages, and then the minimum coverage AOIs for each cluster are determined. Note that there may be overlapping AOIs, in which case the category with the most packages is selected.

    \item Louvain \cite{blondel2008fast}, which converts the AOI segmentation into a community detection problem \cite{blondel2008fast, fortunato2010community}, and applies a classical algorithm called Louvain in the community detection to segment AOI. The method consists of two iterative phases. In the first phase, each node is assigned to its own community. In the second phase, the algorithm tries to maximize modularity gain by merging communities. The two steps are repeated until there is no possible modularity-increasing reassignment of communities.

    \item GCLP \cite{chen2016dynamic}, which formulates the problem as a community detection problem and employs the label propagation method for clustering. It incorporates geographic constraints into the objective function to restrict the size and shape of the clusters. We derive the edge weight matrix based on the trajectory transfer graph. Then, apply GCLP to divide AOIs.
    
\end{itemize}

\subsection{Results in Synthetic Dataset}

We conduct experiments in three different grid sizes: $5\times5$, $6\times6$, and $10\times10$. Based on the results presented in Table~\ref{tab:result}, TrajRL4AOI stands out as the top-performing method across all grid sizes, in both the FMI and Co-AOI rates. This demonstrates the model's robustness and efficacy in AOI segmentation tasks. 

\par For the two basic methods, RoadNetwork demonstrates its effectiveness in dividing the AOI, indicating the rationality of road segmentation. However, due to the complexity of road conditions, optimal performance cannot be achieved. The GreedySeg demonstrates varying performance across grid sizes. Its FMI and Co-AOI rate is relatively low, meaning that there are numerous and small partitions, i.e., fragmented partitioning.

\par Louvain, with its intricate approach to AOI segmentation as a community detection problem, exhibits variability in its results. Its FMI and Co-AOI rate decreases with increasing grid size, hinting at potential scalability issues.
DBSCAN, a clustering method based on parcel locations, yields suboptimal segmentation results as it only considers parcel information. Its FMI rate is not high, indicating misclassifications. Its Co-AOI rate is similar to Louvain, suggesting a relatively fragmented partitioning.

CKMeans incorporates distance constraints to cluster parcels, limiting the occurrence of irregular-shaped partitions. However, considering only geographic locations results in a lack of semantic information in the partitions, leading to a generally lower FMI rate.
GCLP utilizes trajectory transition graphs for community partitioning, combining geographic locations and trajectory information. However, due to the significant impact of the distance function on the value function, the partitioning becomes overly fragmented, resulting in a very low FMI and CR-AOI rate. Benefited by considerations of various rewards, TrajRL4AOI can well balance trajectory switching and road network similarity, and obtain better partitioning. In summary, among the evaluated methods, TrajRL4AOI has proven to be the most reliable and flexible model for accomplishing the task at hand.

\subsection{Effect of Different Parameters}
\par We further investigate the effect of different parameters, including the traversal number of the agent and the reward weight.

\begin{figure}[htbp]
    \centering
    \vspace{-1.5em}
    \includegraphics[width=0.8\linewidth]{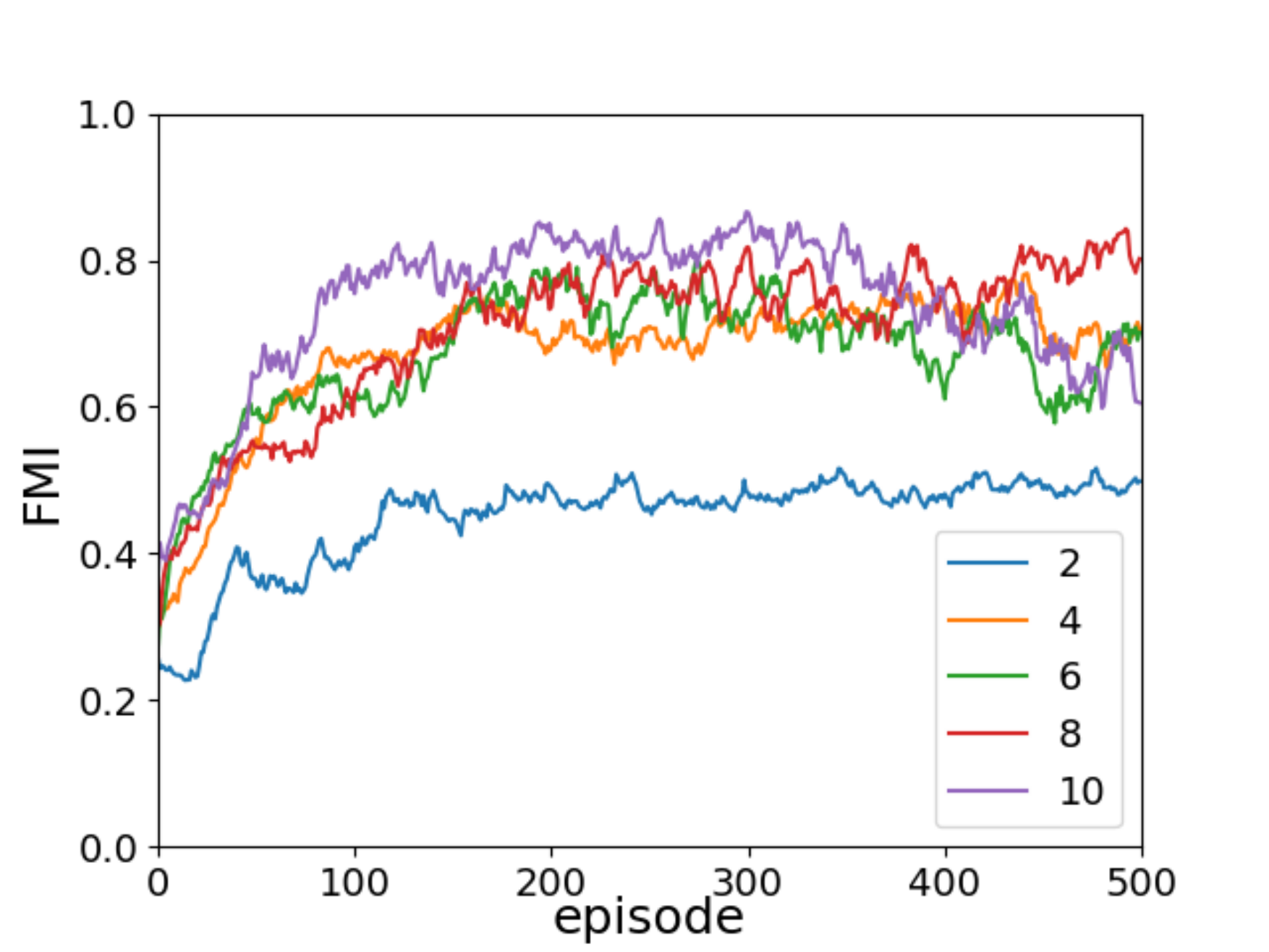}
    \caption{FMI curves under different traversal numbers. When the traversal number is larger than 2, final FMIs are similar. While FMI is lower than others clearly when traversing twice.}
    \label{fig:traversal number}
    \vspace{-0.5em}
\end{figure}

\par \noindent \textbf{The influence of the traversal number.}
 We test the effect of the traversal number on the $10 \times 10$. Figure \ref{fig:traversal number} demonstrates the variation of the partitioning results' FMI with the number of training episodes. Increasing the traversal number beyond 2 has minimal impact on the intelligent agent's partitioning results. This can be attributed to the fact that the agent has sufficient steps to achieve the correct partitioning. However, when the number is limited, the training curve exhibits a noticeable decline. Therefore, the traversal number needs to exceed a certain threshold. To ensure the exploration of correct results and maintain training efficiency, we selected 8 as the optimal number of traversing.

\begin{figure}[htbp]
    \centering
    \vspace{-1em}
    \includegraphics[width=1\linewidth]{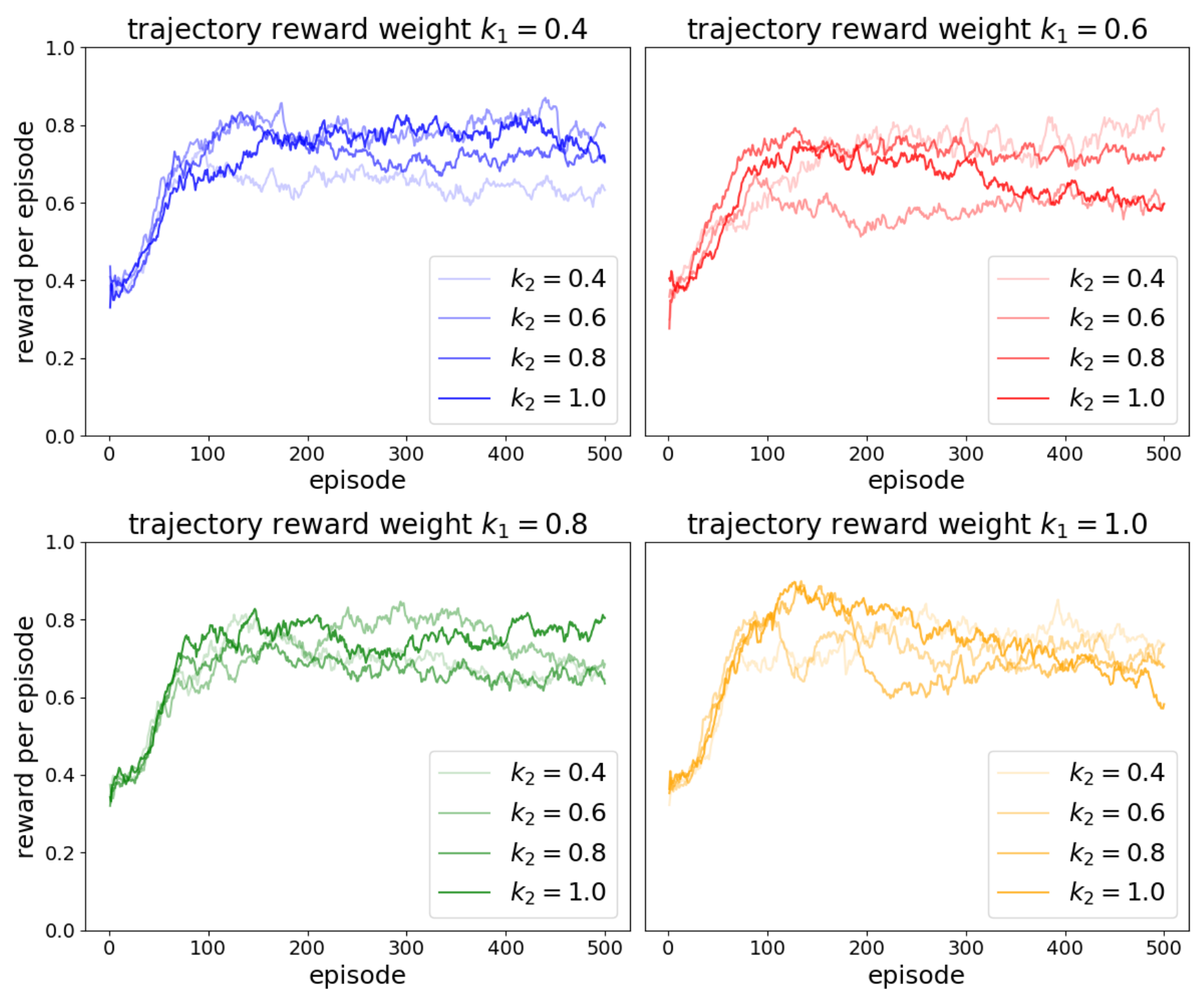}
    \vspace{-1.5em}
    \caption{The FMI curves under different weights. The FMI is maximized when the road reward weight and trajectory network reward weight are set to 0.6:0.4.}
    \label{fig:parameter_effective_reward}
    \vspace{-0.5em}
\end{figure}

\begin{figure*}[!t]
    \centering
    \vspace{-1em}
    \includegraphics[width=1\textwidth]{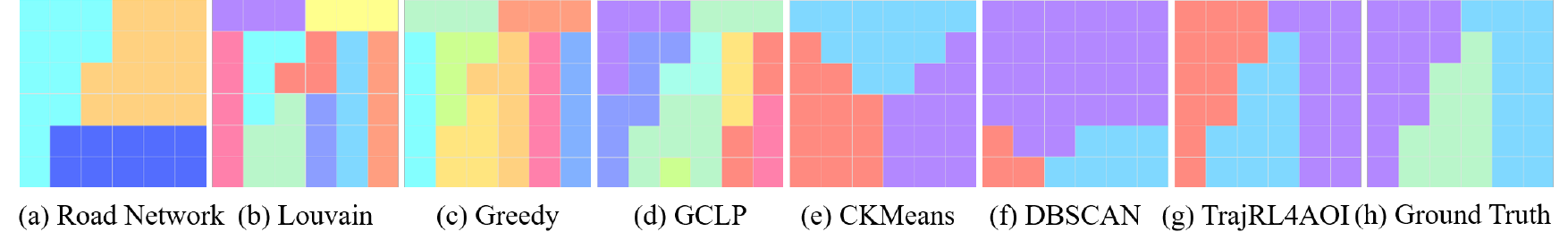}
    \vspace{-2em}
    \caption{ AOI segmentation results of different methods for the $6 \times 6$ AOI environment. Because TrajRL4AOI integrates various input information, the model avoids fragmented partitioning and exhibits the best performance.}
    \label{fig:case_study}
    \vspace{-1em}
\end{figure*}

\begin{figure*}[!t]
    \vspace{1em}
    \centering
    \includegraphics[width=0.95\textwidth]{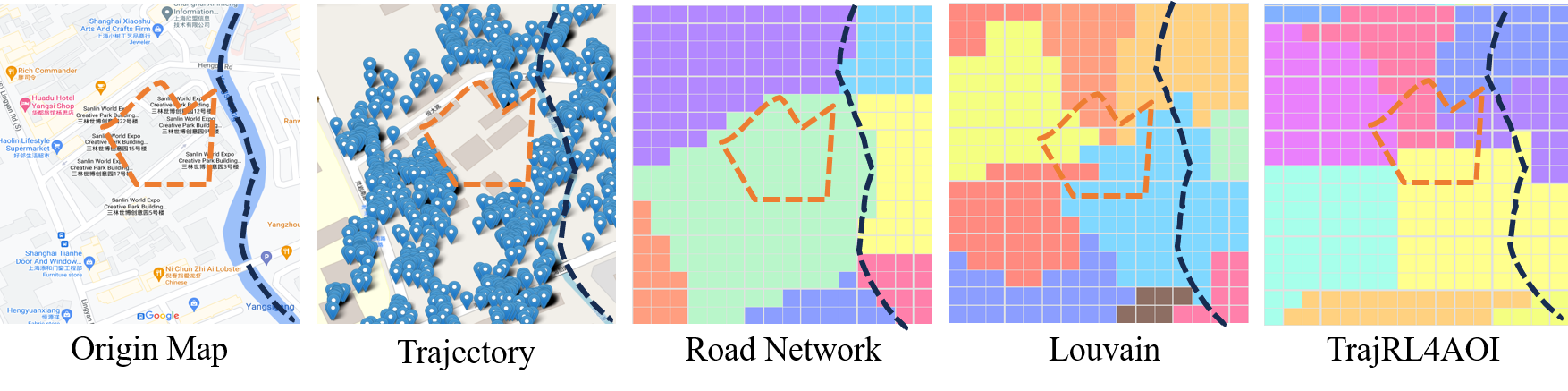}
    \vspace{-1em}
    \caption{AOI segmentation results in real-world data. TrajRL4AOI takes into account both trajectory and road network information. It effectively divides areas with residential zones or fences and also merges two regions separated by a river based on trajectory data, which could address the scenario depicted in Figure \ref{fig:traj_motivation} effectively. }
    \label{fig:real world case_study}
    \vspace{-1em}
\end{figure*}

\par \noindent \textbf{The influence of the reward weight.}
We adjust the weight of both rewards in $10\times10$ and found that the FMI is maximized when the road reward weight and trajectory network reward weight are set to 0.6:0.4. Figure \ref{fig:parameter_effective_reward} illustrates the variation of the FMI with respect to the training episodes under different reward weight parameters. As the training progresses, the FMI consistently rises and gradually converges around 0.8. When the trajectory reward weight is increased, the curve exhibits greater fluctuations, indicating higher training difficulty. And the road network reward has a relatively smaller impact on the training process than the trajectory reward has. This can be attributed to the fact that, compared to the trajectory reward, the road network reward is generally smaller in training.
    


\subsection{Ablation Study}



\par
\begin{figure}
    \centering
    \includegraphics[width=0.8\linewidth]{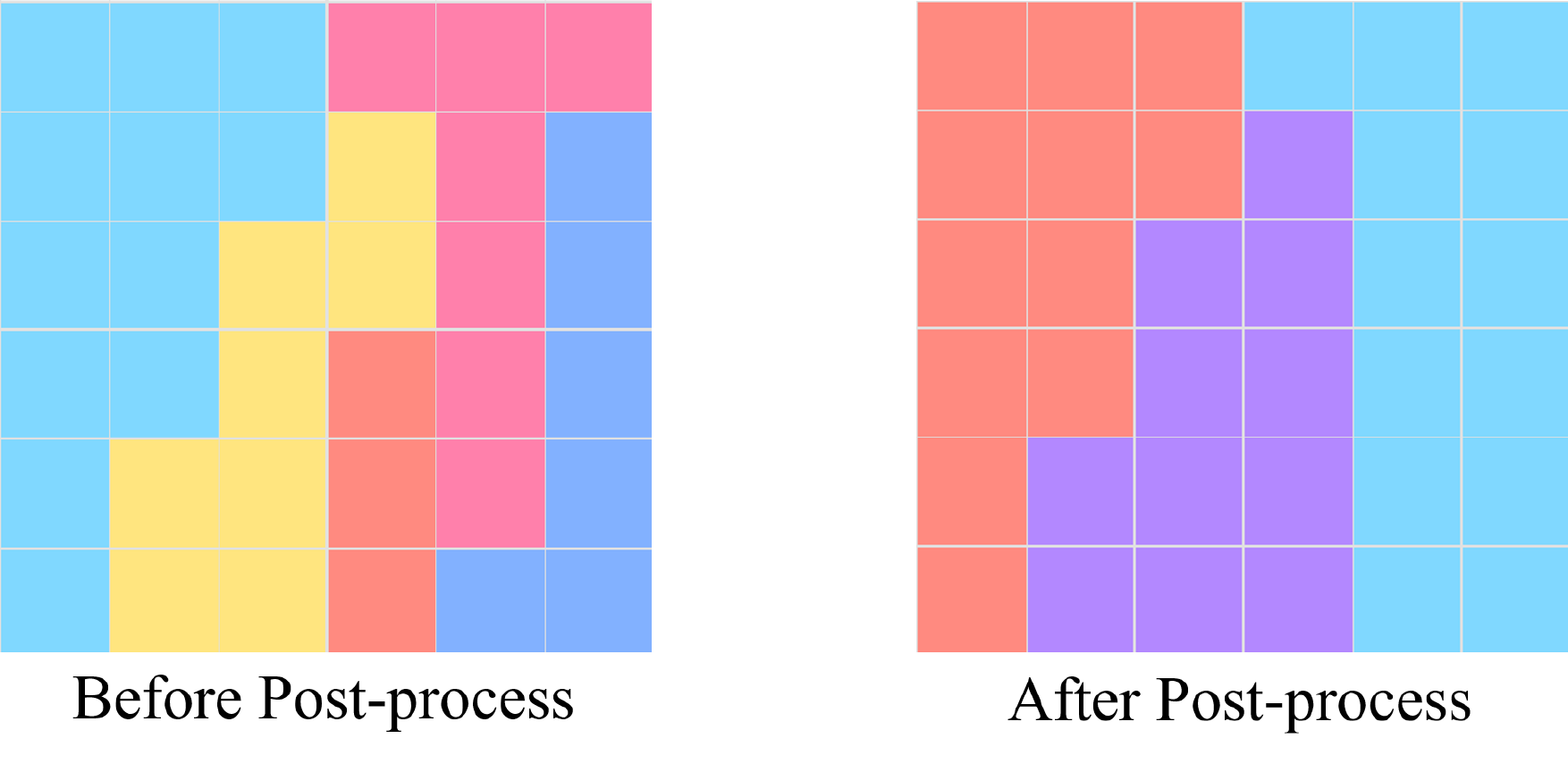}
    \vspace{-1em}
    \caption{Comparison of results before and after post-processing. The post-processing module effectively aggregates AOI regions.}
    \label{fig:post-process ablation}
    \vspace{-1.5em}
\end{figure}

 \noindent \textbf{Post-process.} To validate the aggregation effect of the post-processing, we conduct tests on the final results with and without the post-processing module in $6\times6$ grids. Figure \ref{fig:post-process ablation} shows that the post-processing module effectively aggregates fragmented AOI regions. Before post-processing, the FMI is 81.8\%, and the Co-AOI rate is 69.6\%. After the post-processing, the results aligned with the ground truth partition, resulting in an FMI of 100\% and a Co-AOI rate of 100\%. The FMI increases by 22.2\%, and the Co-AOI rate improves by 43.6\%.

\par \noindent \textbf{Reward.} To assess reward weight impact, we set the trajectory reward weight and road network reward weight to 0, respectively. To make a better comparison, we report the performance of different variants without the post-process (which will change the AOI distribution after the operation). As illustrated in Table \ref{tab:Rewards ablation}, removing either of the rewards decreased model performance. In the optimal setting, for the $6\times6$ synthetic dataset, when the road network reward is omitted, there was a decrease of 17.28\% in FMI and 27.82\% in CR. Similarly, when the trajectory reward is omitted, there was a decrease of 39.82\% in FMI and 54.30\% in CR. Since trajectories can partially represent road information, the results obtained with only the trajectory reward might be more aligned with the road network compared to the results obtained with only the road network reward. Among the two rewards, the trajectory reward had a larger impact on the results. This can be attributed to the fact that the correct partitioning is based on minimizing the number of trajectory switches, and the trajectory reward is better aligned with this objective, hence exerting a greater influence on the results.

\begin{table}
    \centering
    \caption{Results in Reward Ablation.}
    \begin{tabular}{ccccc}
        \toprule
        $6\times6$ & $R_1$ & $R_2$ &FMI& CR \\
        \midrule
        Road (Only) & -17.0 & 32.7\% & 49.2\% & 31.8\% \\
        Trajectory (Only) & -8.4 & 35.9\% & 67.6\% & 50.2\% \\
        Trajectory+Road Reward & -6.1 & 53.6\% & 81.8\% & 69.6\% \\
        \bottomrule
    \end{tabular}
    \label{tab:Rewards ablation}
    \vspace{-1.5em}
\end{table}


\subsection{Case Study}

\par \noindent \textbf{Synthetic data.} On the 6$\times$6 synthetic data, we visualize the partitioning results of baseline methods and TrajRL4AOI. In Figure \ref{fig:case_study}, each color represents one AOI. Figure \ref{fig:case_study}(a) shows the assumed road network information. This partition is similar to the road-network-based Segmentation in Figure \ref{fig:traj_motivation}, which is affected by fences, alleys, etc, thus giving coarse results. Figures \ref{fig:case_study}(b), (c) and (d) present the results based on Louvain, greedy and GCLP algorithms. It can be easily observed that they share similar characteristics: the partitions are too fragmented and fail to merge regions that should be combined. This is because although Louvain and GCLP discover AOIs with better objective functions, they fundamentally start from localized grids. As a result, like the greedy method, they separate regions with lower edge weights than adjacent ones, which naturally leads to many fragmented areas. This phenomenon is exacerbated for GCLP due to the distance loss in its objective function. For CKMeans and DBSCAN, as shown in Figure \ref{fig:case_study}(e) and (f), considering only the parcel information leads to seriously wrong partitioning results. CKMeans produces average partitioning effects as it clusters based on mean distances, while DBSCAN gives the first AOI occupying almost all grids, owing to its iterative nature. In comparison, as shown in Figure \ref{fig:case_study}(g), our method integrates various input information: the road network prior accelerates initial training, trajectory information avoids drawbacks of the road network, etc. Moreover, through multiple rounds of reinforcement learning, Q-learning can focus on the global context. Thus, the same results as GroundTruth (Figure \ref{fig:case_study}(h)) can be achieved.

\par \noindent \textbf{Real-world data.}
We tested our model in the area bounded by $121.4910^{\circ} E - 121.4938^{\circ} E$, $31.1577^{\circ} N - 31.1606^{\circ} N$.

As shown in Figure \ref{fig:real world case_study}(a), the region could be divided into seven areas by three roads and a river. Figure \ref{fig:real world case_study}(c) is the result of the road network-based partitioning, which can effectively identify the separations between areas and allocate them accordingly. However, in the upper right corner of the area, even though there is a river separating it, there are still many trajectories crossing the river, indicating that couriers frequently pass through the river during deliveries. Therefore, these two parts should be considered as the same AOI. The road network-based approach lacks the extraction of trajectory semantics, preventing it from achieving optimal partitioning.

Figure \ref{fig:real world case_study}(d) shows the result by Louvain. It tends to follow the trajectories but still lacks a comprehensive understanding of spatial semantic information, resulting in fragmented AOIs.
In contrast, the partitioning results of TrajRL4AOI effectively avoid overly fragmented divisions in Figure \ref{fig:real world case_study}(e). For small areas in the center of the area, where fewer trajectories pass through, the model accurately captures this information and assigns them to different AOIs. Additionally, the model successfully merges two areas on both sides for the river in the upper right corner, achieving a better partitioning result.

\section{Practicality}

\par
\begin{figure}
    \centering
    \includegraphics[width=1\linewidth]{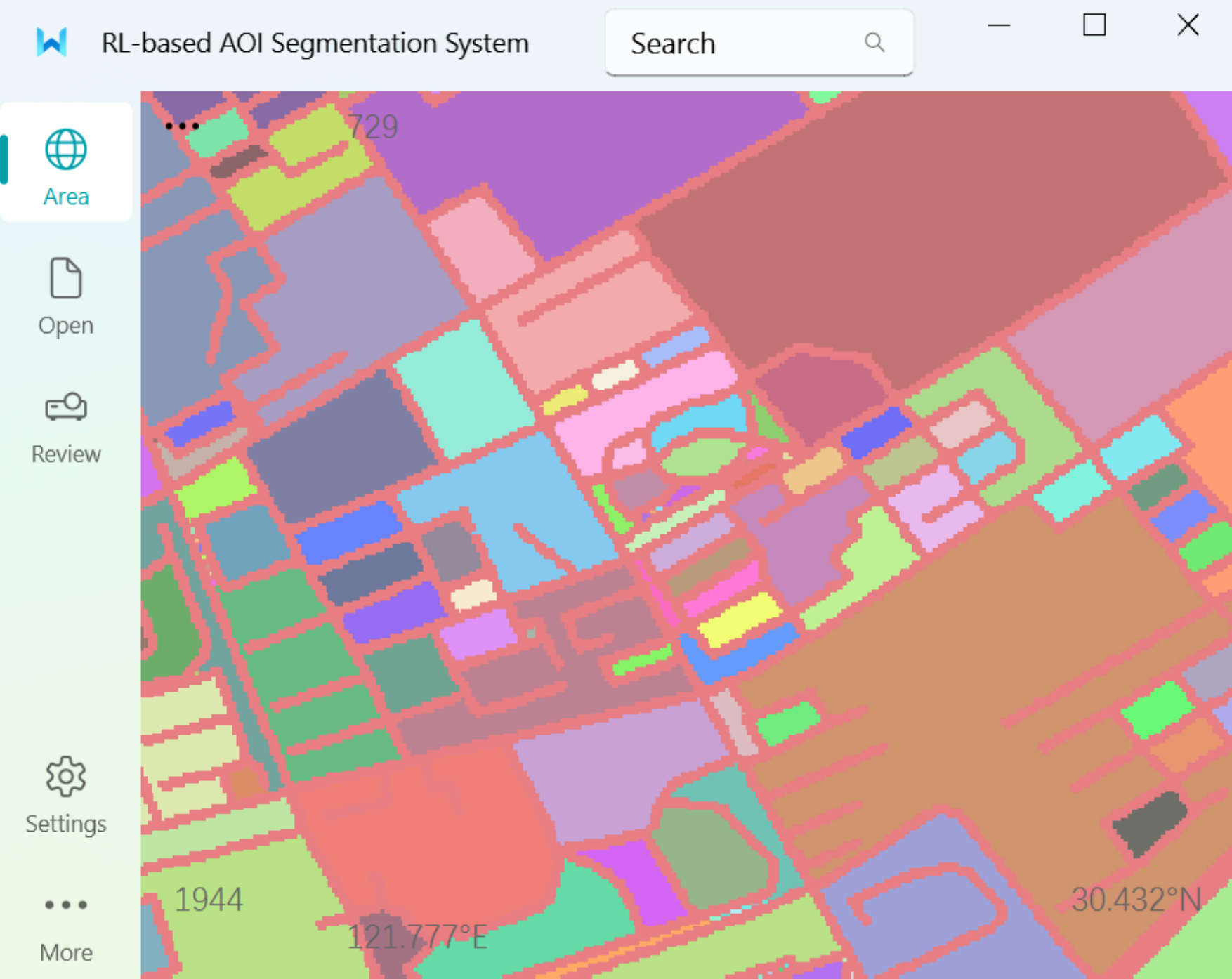}
    \vspace{-1.5em}
    \caption{Reinforement Learning-based AOI Segmentation System. It is a high-performance visualization solution that supports dynamic rendering for AOI grids, parcels, trajectories, and road networks.}
    \label{fig:system}
    \vspace{-1.5em}
\end{figure}

 We propose a high-performance visualization solution. As shown in Figure \ref{fig:system}, we have independently designed and developed a visualization platform that can dynamically render AOI grids, parcels, trajectories, and road networks. It is a city AOI division visualization and display system based on the Qt framework\cite{summerfield2007pyqt}, serving as a visualization platform for AOI division model training, testing and related data display. When used as standalone software, users can drag and drop their own data into the software for automatic loading and rendering. Users can then zoom and pan the map data and view relevant information by clicking or selecting specific areas, such as the coordinates and values of selected areas or elements. The platform can be imported as a Python module into the code for training or testing models, allowing the synchronous display of model process outputs without blocking other code, making it convenient for developers to monitor model status. The platform also supports the temporary storage, recording, and replaying of process data, allowing users to record certain data states for repeated study or sharing with others. The platform's architectural design offers high flexibility and scalability, enabling users to customize their dedicated visualization platform by modifying a few configuration files. This greatly enhances research efficiency as it supports user-defined events and rendering layers tailored to their specific needs. In summary, the platform revolves around layer objects and supports asynchronous rendering, event-driven operations, and state recording. It has good scalability and robustness, and we believe it will continue to provide strong support for future related work. 

\section{Conclusion}


\par In this paper, we addressed the limitation of traditional AOI segmentation algorithms in Location-Based Services, which primarily depend on road networks and lack semantic information. We introduced a novel DRL-based framework, DRL4ROI, which harnesses Deep Reinforcement Learning for AOI segmentation. This framework employs semantic information as the reward to guide AOI generation, ensuring better efficiency and service quality. We further developed a model, TrajRL4AOI, which emphasizes trajectory modularity and road network matchness for enhanced AOI segmentation in logistics platforms. Utilizing a Double Deep Q-learning Network, the model refines the AOI generation during training. Empirical tests on the synthetic dataset and real-world dataset confirmed the effectiveness and superiority of our proposed methodology. 


\par In future work, we are considering introducing more abundant information, such as satellite images in the proposed framework, for more high-quality AOI segmentation.

\section*{ACKNOWLEDGMENT}
This work was supported by the National Natural Science Foundation of China (No. 62372031). Data is supported by cainiao.


\newpage

\bibliographystyle{IEEEtran}
\bibliography{my_bib}

\newpage

\end{document}